\documentclass[10pt,twocolumn,letterpaper]{article}
\usepackage[belowskip=-10pt,aboveskip=10pt]{caption}

\usepackage{svg}
\usepackage{booktabs}

\usepackage[pagenumbers]{cvpr} % To force page numbers, e.g. for an arXiv version

\definecolor{cvprblue}{rgb}{0.21,0.49,0.74}
\usepackage[pagebackref,breaklinks,colorlinks,allcolors=cvprblue]{hyperref}

\def\paperID{7} % *** Enter the Paper ID here
\def\confName{CVPR}
\def\confYear{2025}

\title{Test Automation for Interactive Scenarios via Promptable Traffic Simulation}

\author{Augusto Mondelli$^*$ \quad Yueshan Li$^*$ \quad Alessandro Zanardi \quad Emilio Frazzoli\\
ETH Zurich\\
{\tt\small yuesli@ethz.ch}
}

\begin{document}
\maketitle
\def\thefootnote{*}\footnotetext{These authors contributed equally to this work}

\begin{abstract}
Autonomous vehicle (AV) planners must undergo rigorous evaluation before widespread deployment on public roads, particularly to assess their robustness against the uncertainty of human behaviors. While recent advancements in data-driven scenario generation enable the simulation of realistic human behaviors in interactive settings, leveraging these models to construct comprehensive tests for AV planners remains an open challenge. In this work, we introduce an automated method to efficiently generate realistic and safety-critical human behaviors for AV planner evaluation in interactive scenarios. We parameterize complex human behaviors using low-dimensional goal positions, which are then fed into a promptable traffic simulator, ProSim, to guide the behaviors of simulated agents. To automate test generation, we introduce a prompt generation module that explores the goal domain and efficiently identifies safety-critical behaviors using Bayesian optimization. We apply our method to the evaluation of an optimization-based planner and demonstrate its effectiveness and efficiency in automatically generating diverse and realistic driving behaviors across scenarios with varying initial conditions.
\end{abstract}    
\section{Introduction}
\label{sec:intro}

\begin{figure}
\includegraphics[width=8.2cm]{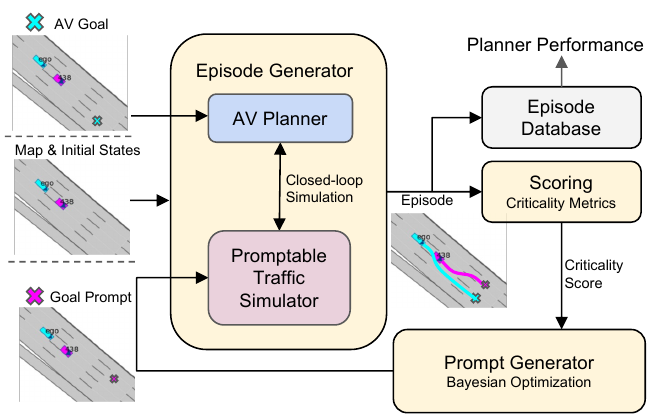}
\caption{\textbf{Overview of the proposed method for test automation in interactive scenarios}. The planner under test runs in closed-loop with the simulated agents, whose behaviors are generated by a promptable, data-driven traffic simulator. By sampling the goal prompts with Bayesian optimization, we create realistic and safety-critical test cases in a automated manner.}
\label{fig:overview}
\vspace{-5pt}
\end{figure}
Autonomous vehicles (AVs) must be capable of handling diverse behaviors of human drivers in interactive scenarios. To ensure the reliability of AV planners, they must undergo comprehensive evaluation before being deployed on public roads together with human drivers.

\textbf{Planner Evaluation.}
Real-world road testing has been widely employed to assess the safety performance of AV planners \cite{di2024comparative}. However, most of the recorded driving logs lack interactivity \cite{ding2025surprise}, limiting the reliability of such evaluations in assessing planner robustness against diverse human behaviors. Scenario-based testing \cite{sun2021scenario} offers a more effective alternative due to its flexibility to target rare and safety-critical interactive behaviors. Nevertheless, within a given scenario, the infinite number of possible driving behaviors presents a significant challenge in identifying a small yet representative set of test cases to draw reliable conclusions regarding the robustness of an AV planner. Search-based methods \cite{ding2021multimodal, huang2024bayesian} have been employed to automatically generate safety-critical interactive scenarios. However, these approaches use simplified models to describe human behaviors, limiting the realism of the generated test cases.

\textbf{Data-driven Traffic Simulation.}
The emergence of data-driven traffic simulators offers great potential to advance test automation in interactive scenarios. In particular, adversarial generation \cite{rempe2022generating, zhong2023guided, chang2024safe} deliberately perturbs the behaviors of surrounding agents to attack the AV planner. These methods can efficiently identify safety-critical behaviors, but suffer from compromised realism due to their adversarial nature. Moreover, the adversarial behaviors are only controllable through embedded objective functions, which requires expert knowledge to design, thereby limiting the broader applicability of these methods. With advancements in large language models (LLMs), the language-conditioned generation \cite{zhong2023language, tan2023language, zhang2024chatscene, xia2025language} has emerged, enabling direct and intuitive control over scenarios. While this approach is able to generate realistic interactions following the language descriptions, it remains unclear how to automatically create these textual descriptions to thoroughly evaluate the robustness of an AV planner. In \cite{tan2024promptable}, controllability of language-conditioned generation is expanded to multimodal prompts, including numerical goal states, which facilitates automated behavior generation in interactive scenarios.

% \subsubsection*{Contribution}
\textbf{Contribution.}
In this work, we propose an automated method for stress-testing the safety of AV planners in interactive scenarios. At the core of our approach is the efficient utilization of a promptable traffic simulator, ProSim \cite{tan2024promptable}, as illustrated in \cref{fig:overview}. We introduce a prompt generation module to automate test case generation by sampling numerical goal position prompts within a user-configured goal domain. To further enhance the testing efficiency, we integrate Bayesian optimization (BO) to identify the goal positions that induce safety-critical human behaviors. Unlike prior methods \cite{gangopadhyay2019identification, huang2024bayesian}, which struggle with the high dimensionality of the scenario spaces, we parameterize realistic human behaviors using low-dimensional numerical prompts, leveraging the expressiveness of the data-driven traffic simulator. As a result, our method can generate realistic and diverse test cases and efficiently identify the safety-critical ones with a small number of samples.

We apply the proposed method to evaluate a model-predictive-control-based planner. We demonstrate the effectiveness of our approach in generating diverse behaviors within a 2-agent scenario across various initial conditions and we analyze the efficiency of BO in identifying safety critical driving behaviors.

In summary, we make the following contributions:
\begin{itemize}
    \item We introduce a novel approach for automated stress-testing of AV planners in interactive scenarios. By sampling goal position prompts and giving them as input to a data-driven traffic simulator, our method generates diverse and realistic human behaviors that interact with the AV planner in closed-loop.
    \item We incorporate Bayesian optimization to efficiently identify safety-critical behaviors. By parameterizing complex human behaviors into a low-dimensional goal domain, our method achieves high sampling efficiency while preserving behavioral realism.
    \item Our method makes no prior assumptions about the planner under test. We demonstrate its effectiveness and efficiency using an external rule-based planner across scenarios with varying initial conditions.
\end{itemize}

\section{Methods}
\textbf{Problem Formulation}
We consider a traffic scenario with one AV governed by the planner under test, referred to as the \textit{ego agent}, and $N$ \textit{simulated agents}. Each agent aims to reach its own goal position, without knowing the goals of others. This incomplete information is a key source of driving hazards, causing agents to commit to incorrect beliefs and engage in safety-critical behaviors. 

In this work, we focus on stress-testing the planner under uncertainty arising from the diverse and unknown goal positions of other agents, while keeping the map, initial states, and the goal of ego agent fixed. In the closed-loop simulation, both the ego agent and simulated agents plan a sequence of future actions to approach their goals, based on the current state observations of all agents in the scene. As the simulation progresses, agents update their observations, refine their plans, and adjust their behavior to avoid collisions. To comprehensively evaluate the robustness of the planner in the interactive scenario, our method generates a collection of episodes, each capturing a realization of the interactive trajectories of the agents. These episodes serve as structured test cases for assessing the AV planner’s safety performance against diverse human behaviors.

% We denote the states of the ego agent at time $t$ as $x_t^E$ and the state of simulated agent $i$ as $x^{i}_t$. We use $x^{-E}_t$ to represent the states of all simulated agents at time $t$. The duration of the simulation is denoted by $T$ and the collection of states from time $t_1$ to $t_2$ is denoted by $x_{t_1:t_2}$. We define a trajectory as a sequence of states from $t=0$ to $T$, denoted by $\tau := x_{0:T}$. The goal positions of the ego and simulated agents are denoted by $G^E$ and $G^{-E}$, respectively. Note that due to the potential interactions, the trajectories may not end at the corresponding goal positions. We then define an \textit{episode} $e$ as a realization of the possible interaction, which contains the trajectories of all agents under a goal assignment of the simulated agents, i.e. $e(G^{-E}) := <\tau^{-E}, \tau^E>$.

\begin{figure}
    \includegraphics[width=8.2cm]{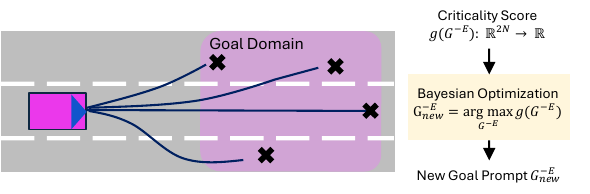}
    \caption{\textbf{The prompt generation module.} This module applies BO to sample the next goal position, aiming to balance between the exploration of the unvisited goal domain and the exploitation of the critical goal assignments. The black crosses represent the sampled goal positions and the corresponding line of each cross demonstrates the simulated trajectory approaching that goal.}
    \label{fig:prompt}
  \end{figure}

\textbf{Prompt Generation Module.}
The prompt generation module samples the goal position prompts and feeds them to the promptable traffic simulator to generate diverse, interactive human behaviors. As shown in \cref{fig:prompt}, it employs Bayesian optimization to explore the goal domain and efficiently identify the prompts that lead to safety-critical human behaviors.

There are several advantages in parameterizing human behaviors with the goal prompts. First, it provides a simple and intuitive way to control generated human behaviors without sacrificing their realism. Secondly, thanks to the low-dimensionality of the goal prompts, it allows Bayesian optimization to identify the safety-critical goal prompts with a small number of samples and ensures its scalability regarding the number of agents. Moreover, it permits user-defined goal domain to constrain the scope of human behaviors that one wants to test against, from traffic-rule-compliant behaviors to the extreme case of all physically permitted behaviors. Such scope requirement can be specified using the reachable set of an agent, which can be efficiently estimated using existing tools \cite{bansal2021deepreach, liu2022commonroad}. 

To run BO, we use a Gaussian process with Matern Kernel as the surrogate model and Upper Confidence Bound as the acquisition function. We choose the goal domain that covers all drivable area in front of the simulated agents.
  
\begin{figure}
    \includegraphics[width=8.2cm]{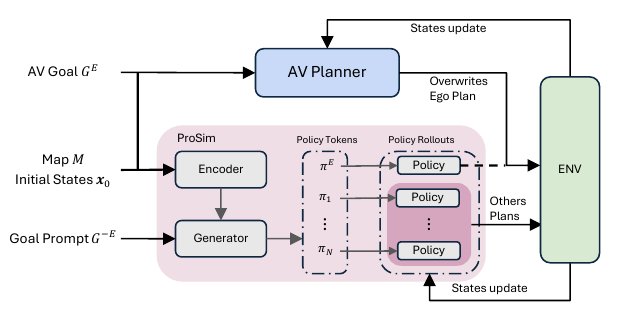}
    \caption{\textbf{The episode generation module.} We integrate an external AV planner in closed-loop with ProSim. Inside ProSim, a policy token is generated for each agent once per simulation run, conditioned on the prompts given and the encoded scene. Then at the policy rollout phase, we overwrite the ego plan generated by ProSim with the actual plan from the AV planner. Both the planner and policy rollout are called recursively with the updated state observations of all agents to allow closed-loop simulation. }
    \label{fig:episode}
    \vspace{-5pt}
  \end{figure}

\textbf{Episode Generation Module.}
The aim of this module is to place the AV in an interactive scenario and simulate its behavior against agents with unknown goals. A crucial aspect of the interaction is the ability of all agents to adapt to the behaviors of others. We ensure the realism of these interactions by generating the behaviors of both the AV and other agents in a closed-loop manner, as illustrated in \cref{fig:episode}.

In this work, we explore the use of ProSim \cite{tan2024promptable}, a promptable traffic simulator, to generate realistic behaviors of other agents in closed-loop with an external AV planner. We selected ProSim due to its capability to generate realistic driver behaviors in closed-loop and its flexibility in accepting prompts in various forms, particularly numerical goal state prompts. Although the original work focused on traffic simulation rather than AV planner evaluation, we successfully integrated it with an external planner in the policy roll-out phase. Specifically, the static map and the initial states of all agents are first encoded as a scene token. A transformer-based generator then processes the goal prompts and the scene token to create a policy token for each agent. The policy module subsequently generates the next states of each agent in a recursive manner, conditioned on the policy token of that agent and the current states of all. At each simulation step, the AV planner also receives the current states of all agents and overwrites the simulated states of the ego agent with the future states it plans to go. The updated states of all agents are then fed back to both the AV planner and the simulator, ensuring realistic reactions throughout the interaction.
  
\textbf{Scoring Module.}
This module evaluates the criticality of a generated episode, which guides the prompt generation module to discover safety-critical human behaviors. The scoring is done in hindsight because of the black-box nature of the planner and reactive nature of the simulation--we do not know how exactly the trajectories of other agents will be and how the planner will interact with them. In this work, we use the minimum absolute distance between the AV and other agents to measure the criticality $g$ of the generated behaviors, i.e.
\small
\begin{equation}
    g = -\min_{t\in \{1,...,T\}} \min_{n\in\{1,...,N\}} \lVert x_t^E - x_t^{n}\rVert_2,
\end{equation}
\normalsize
where $x_t^E$ and $x_t^n$ denote the state of the ego agent and the simulated agent $i$ at time $t$, respectively.
This score serves as the objective to be maximized using Bayesian optimization. Notably, our formulation can be adapted to other user-defined metrics as episode scores.
\section{Experiments}
In this section, we first demonstrate the effectiveness of our method in automatically generating realistic, diverse and safety-critical behaviors in 2-agent highway scenarios with varying initial conditions. Then, we conduct quantitative experiments to validate the efficiency of employing BO to identify safety-critical behaviors. 

\textbf{AV Planner under test.}
We apply our method to test a planner based on model predictive control, which solves an optimization to find a sequence of future actions that reach its goal as fast as possible, while avoiding collision with other agents. Since the goals of other agents are unknown to the planner, it predicts the future trajectories of other agents by assuming they always try to track the current lane center with a constant velocity. This assumption will be violated once the other agents are prompted differently, resulting in challenging test cases that reveal the robustness of the planner at interaction.

\textbf{Data-driven Traffic Simulator.} In this experiment, we use the checkpoint released by the authors of ProSim without any fine-tuning. It is worth mentioning that while ProSim is trained on maps from Waymo Open Motion Dataset (WOMD) \cite{ettinger2021large}, the planner under test is originally developed on the CommonRoad \cite{althoff2017commonroad} map configuration. To run closed-loop simulations with the two components, we convert the CommonRoad map to the WOMD configuration before initializing ProSim.

\begin{figure}
  \centering
  % \begin{subfigure}{0.12\linewidth}
    \includegraphics[width=2.5cm]{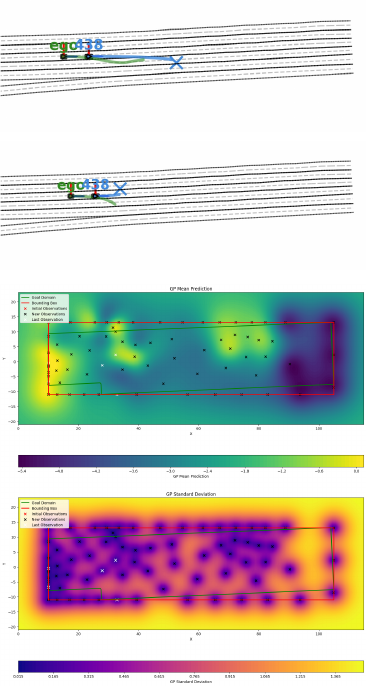}
    % \caption{Scenario 1: The other agent is in front of the ego agent.}
    \label{fig:short-a}
  % \end{subfigure}
  \hfill
  % \begin{subfigure}{0.12\linewidth}
    \includegraphics[width=2.5cm]{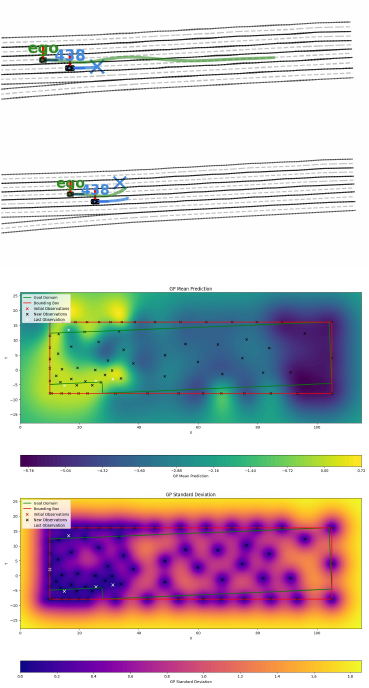}
    % \caption{Scenario 2: The other agent is in the front right of the ego agent.}
    \label{fig:short-a}
  % \end{subfigure}
  \hfill
  % \begin{subfigure}{0.12\linewidth}
    \includegraphics[width=2.5cm]{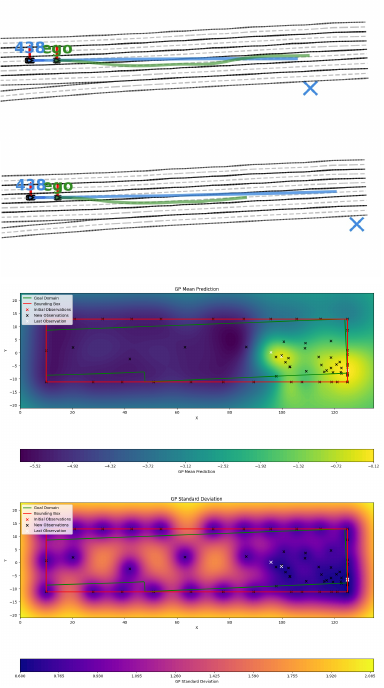}
    % \caption{Scenario 3: The other agent is behind the ego agent.}
    \label{fig:short-b}
  % \end{subfigure}
  \caption{\textbf{Qualitative results of the automated human behavior generation.} The top two rows show examples of the generated episodes. The green and blue lines show the trajectories of the ego agent and the simulated agent, respectively. The blue cross shows the goal prompt. The last two rows show the mean and variance of the fitted Gaussian process within the goal domain. Our method automatically finds diverse safety-critical behaviors specific to each scenario. \textit{Left:}The other agent is in front of the ego agent. \textit{Middle:} The other agent is in the front right of the ego agent. \textit{Right:} The other agent is behind the ego agent.}
  \label{fig:results}
  \vspace{-5pt}
\end{figure}

\textbf{Evaluation of the automated behavior generation.}
We tested our method in 2-agent highway scenarios with three initial positions of the other agent: in front of AV; in the front right; and behind. The goal of the AV is to change to the right lane. In each test, we run BO with 75 iterations without initialization. Examples of the generated episodes and the fitted Gaussian process are shown in \cref{fig:results}. In the first two cases, BO identified that the planner is likely to cause a rear-end collision if the simulated agent brakes. This outcome is intuitive, as the planner initially assumes that the simulated agent will maintain its current velocity, allowing the AV to safely steer into the right lane. However, if this assumption is violated —for instance, in case an animal suddenly appears in front of the simulated agent— the planner does not maintain a sufficient safety margin to react appropriately. In contrast, a very different safety-critical behavior emerges in the third case: the simulated agent attempts to overtake and merge into the right lane ahead of the AV. Since the planner does not account for rear agents and assumes that other vehicles will remain in their lanes, collisions occur as these assumptions are violated. These examples empirically validate the effectiveness of our method in automatically discovering safety-critical behaviors in various initial conditions.

\textbf{Effectiveness of Bayesian optimization}
We compare 75 episodes generated with final goals sampled by BO to 75 other episodes generated with final goals obtained through random sampling of the goal domain using Sobol sampling ("Random"). BO is performed without any initialization of the Gaussian process. We do so for each of the three initial conditions described above.
We measure the safety-criticality and the diversity of the generated episodes. For safety-criticality, we consider the collision rate (``Coll"), the minimum distance between the ego and other agent (``Min Dist") and the Time-to-Collision (``TTC"). For diversity we compute the pairwise average self-distance as in \cite{cui2021lookout} for both the ego (``EgoASD") and the other agent (``AgentASD"), i.e.
\small
\begin{align}
    % \textit{EgoASD} &= \frac{1}{n_e(n_e-1)}\sum_{i=1}^{n_e}\sum_{j=i+1}^{n_e}\lVert \tau_i^E - \tau_j^E \rVert_2, \\
    % \textit{AgentASD} &= \frac{1}{n_e(n_e-1)}\sum_{i=1}^{n_e}\sum_{j=i+1}^{n_e}\lVert \tau_i^{-E} - \tau_j^{-E} \rVert_2,
    \textit{EgoASD} &= \frac{1}{n_e(n_e-1)}\sum_{i=1}^{n_e}\sum_{j=i+1}^{n_e} d( \tau_i^E, \tau_j^E), \\
    \textit{AgentASD} &= \frac{1}{n_e(n_e-1)}\sum_{i=1}^{n_e}\sum_{j=i+1}^{n_e} d(\tau_i^{-E}, \tau_j^{-E}),
\end{align}
\normalsize
with $n_s$ denoting the total number of episodes, $\tau_i^{E}$ and $\tau_i^{-E}$ denoting the trajectories of the ego and the simulated agent of episode $i$, respectively, and $d$ denoting the mean Euclidean distance between corresponding states of the two trajectories, computed over all simulation timesteps. The agent diversity provides a straightforward measurement on how different goal prompts lead to different human behaviors, while the ego diversity captures how much the AV planner has to adapt to the behavior of others due to interaction. For episodes with a collision, we exclude the trajectories after collision in the metrics.

\begin{table}[ht]
% \ra{1.3}
\resizebox{0.46\textwidth}{!}{\begin{tabular}{@{}rrrrcrrr@{}}\toprule
& \multicolumn{3}{c}{Safety-criticality} & \phantom{abc}& \multicolumn{2}{c}{Diversity}\\
\cmidrule{2-4} \cmidrule{6-7} 
& Coll ($\%$)$\uparrow$ & Min Dist (m)$\downarrow$ & TTC (s)$\downarrow$ && EgoASD $\uparrow$ & AgentASD$\uparrow$\\ 
\midrule
Front \\
Random &  $0$& $2.72\pm1.52$ & $0.84\pm0.67$ && $\mathbf{10.86}$ & $11.03$ \\
BO &  $\mathbf{16}$& $\mathbf{1.91\pm1.47}$& $\mathbf{0.53\pm0.72}$&& $10.73$& $\mathbf{12.15}$\\ \midrule
Front right \\
Random &  $12$& $3.01\pm1.66$ & $1.34\pm0.93$ && $\mathbf{10.65}$ & $11.56$ \\
BO &  $\mathbf{35}$& $\mathbf{1.84\pm1.87}$& $\mathbf{0.66\pm0.86}$&& $9.57$& $\mathbf{12.10}$\\ 
\midrule
Behind \\
Random & $3$ & $4.22\pm1.66$ & $1.39\pm0.99$ && $0.39$ & $\mathbf{13.85}$ \\
BO & $\mathbf{21}$ & $\mathbf{1.98\pm2.11}$& $\mathbf{0.63\pm 0.76}$&& $\mathbf{4.48}$& $12.39$\\
\bottomrule
\end{tabular}}
\caption{\textbf{Effectiveness of BO.} With the same number of samples, BO manages to find more safety critical behaviors while achieving comparable diversity as random sampling. 
% \textit{Scenario 1:} The other agent is in front of the ego agent. \textit{Scenario 2:} The other agent is in the front right of the ego agent. \textit{Scenario 3:} The other agent is behind the ego agent.
}
\label{tab:comparison}
\end{table}

 The experiment results are summarized in \cref{tab:comparison}. BO outperforms random sampling in all criticality metrics, showing its efficiency in discovering the safety-critical human behaviors. It also achieves comparable diversity as random sampling, even though the latter are specially designed to generating well-distributed samples.

\section{Conclusions}
In this work, we introduce a novel method to automate human behavior generation in interactive scenarios. By leveraging the expressiveness of a promptable, data-driven traffic simulator, our approach generates diverse and realistic human behaviors by sampling goal prompts in a low-dimensional space. Furthermore, with the integration of Bayesian Optimization, our method efficiently identifies safety-critical behaviors with a small number of samples. Our method paves the way for automated and reliable evaluation of AV planners, harnessing it with the latest advancements in data-driven generative models. 

We note that, since our method is built upon the assumption that the uncertainty of human behaviors arises from their unknown goal positions, other unsafe behaviors due to human irrationality, such as they being drunk, distracted or at panic, cannot be covered by the current form of prompting. Additionally, while data-driven simulators are powerful in generating the most likely human behaviors leveraging the rich traffic datasets, they fall short in creating the out-of-distribution behaviors, which is a non-neglectable cause of traffic accidents. Therefore, we envision a hybrid approach that combines data-driven and behavioral model-based simulation to comprehensively evaluate the safety performance of AV planners.
{
    \small
    \bibliographystyle{ieeenat_fullname}
    \bibliography{main}
}

% WARNING: do not forget to delete the supplementary pages from your submission 
% \input{sec/X_suppl}

\end{document}